\documentclass[conference]{IEEEtran}
\IEEEoverridecommandlockouts

\usepackage{cite}
\usepackage{amsmath,amssymb,amsfonts}
\usepackage{algorithmic}
\usepackage{graphicx}
\usepackage{textcomp}
\usepackage{xcolor}
\usepackage{wrapfig}
\usepackage{booktabs} 
\usepackage{multirow}
\usepackage[colorlinks,linkcolor=red]{hyperref}

\def\BibTeX{{\rm B\kern-.05em{\sc i\kern-.025em b}\kern-.08em
    T\kern-.1667em\lower.7ex\hbox{E}\kern-.125emX}}
\begin{document}

\title{Gaussian Difference: Find Any Change Instance in 3D Scenes\\

\author{
    \IEEEauthorblockN{
        Binbin Jiang\IEEEauthorrefmark{1}, 
        Rui Huang\IEEEauthorrefmark{1}, 
        Qingyi Zhao\IEEEauthorrefmark{1}, 
        Yuxiang Zhang\IEEEauthorrefmark{1}
    }
    \IEEEauthorblockA{
        \IEEEauthorrefmark{1}College of Computer Science and Technology, Civil Aviation University of China, Tianjin, P.R. China\\
        Email: 
            \{2022051012, rhuang, 2022051007, yxzhang\}@cauc.edu.cn, 
    }
}


}
\maketitle

\begin{abstract}

Instance-level change detection in 3D scenes presents significant challenges, particularly in uncontrolled environments lacking labeled image pairs, consistent camera poses, or uniform lighting conditions. This paper addresses these challenges by introducing a novel approach for detecting changes in real-world scenarios. Our method leverages 4D Gaussians to embed multiple images into Gaussian distributions, enabling the rendering of two coherent image sequences. We segment each image and assign unique identifiers to instances, facilitating efficient change detection through ID comparison. Additionally, we utilize change maps and classification encodings to categorize 4D Gaussians as changed or unchanged, allowing for the rendering of comprehensive change maps from any viewpoint. Extensive experiments across various instance-level change detection datasets demonstrate that our method significantly outperforms state-of-the-art approaches like C-NERF and CYWS-3D, especially in scenarios with substantial lighting variations. Our approach offers improved detection accuracy, robustness to lighting changes, and efficient processing times, advancing the field of 3D change detection.

\end{abstract}


\section{Introduction}

Most existing change detection (CD) methods rely on coupled images and identify changes based on 2D appearance features, These approaches have been extensively researched across various domains, including street-view scenes \cite{ sakurada2013detecting, alcantarilla2018street} and remote sensing \cite{2d_captioning_jhamtani2018learning, chen2020spatial, huang2023background}.  Lei et al. \cite{lei2020hierarchical} propose a hierarchical paired channel fusion network to improve the accuracy of change maps.
Wang et al. \cite{wang2023cross} propose an effective multi-level feature interaction method to achieve high-performance change detection. Despite these improvements, such CD methods face significant limitations. They depend heavily on large-scale labeled datasets (e.g., \cite{chen2020spatial,ji2018fully, lebedev2018change}) that require precise camera pose alignment, which can be a significant challenge.

In contrast to conventional CD methods, 3D CD aims to identify changes in a scene over time by leveraging the geometric information inherent in 3D representations. A key advantage of 3D CD is its ability to generate change maps from arbitrary viewpoints \cite{ulusoy2014image,6942806,qin20163d,huang2023c}, offering greater flexibility and robustness compared to traditional 2D methods. Typically, 3D CD utilizes a combination of multi-view images and depth information captured both before and after scene changes, providing a more comprehensive basis for change analysis.
%
%
%
%
Fehr et al.\cite{7989614}  develop a 3D reconstruction algorithm that utilizes an extended Truncated Signed Distance Function to solve the scene differencing problem more accurately. 
Ku et al.\cite{ku2021shrec} contribute a street-scene dataset for 3D point cloud change detection that can detect changes in a complex street environment from multi-temporal point clouds. 
Qiu et al.\cite{qiu20233d} propose a method to explicitly localize changes in 3D bounding boxes from two point clouds. 
Sachdeva et al.\cite{sachdeva2023change2D} propose a method for change detection by using depth maps to map the images before and after changes into point clouds. Achieving high-quality 3D point clouds, especially from real-world scenes, can indeed be quite expensive and time-consuming.
Recent research \cite{huang2023c} has proposed a CD method, \textsc{C-NeRF}, grounded in Neural Radiance Fields (NeRF) \cite{mildenhall2020nerf}, which enables change mapping with only two image sequences captured from the scene before and after the change occurs.
\textsc{C-NeRF}  constructs two aligned NeRFs and renders image pairs to calculate absolute differences for change identification. While C-NERF offers the advantage of rendering high-quality change maps from novel viewpoints, it faces limitations in scenes with substantial lighting variations and requires considerable processing time.
%
%

\begin{figure*}[!ht]
  \centering
  \includegraphics[width=\linewidth]{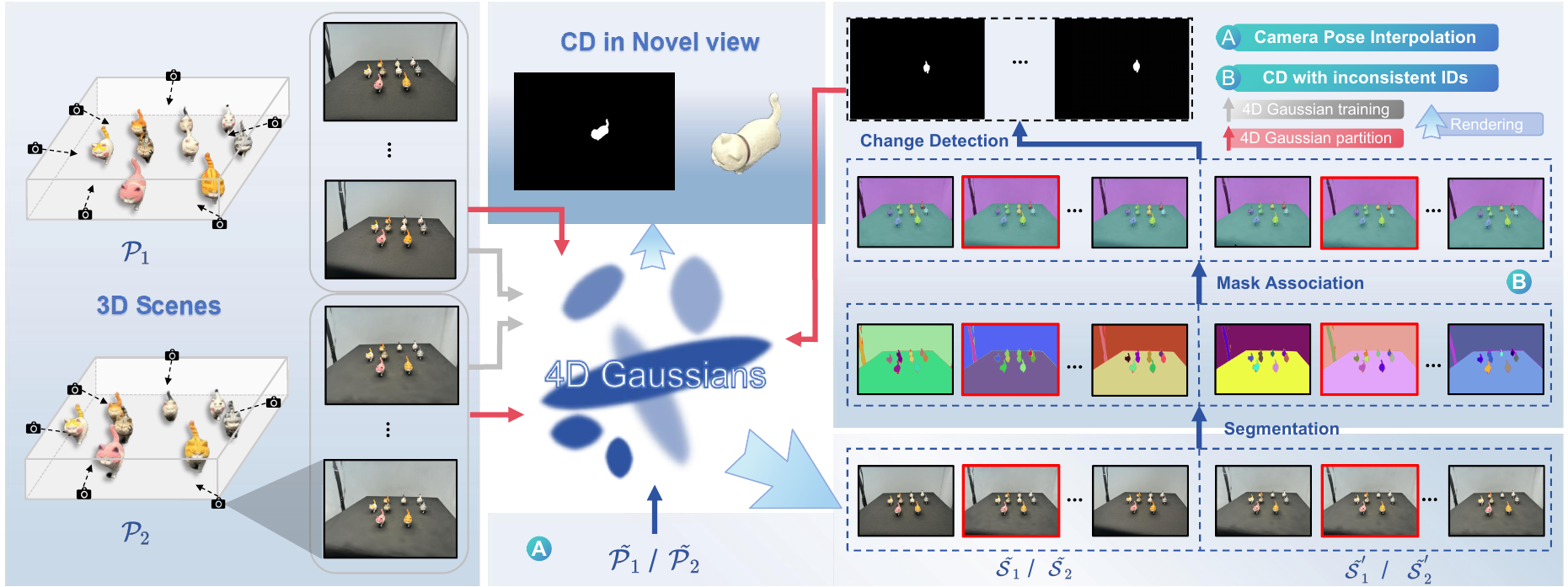}
  \caption{The main framework of the proposed 3D change detection method.} 
  \label{fig:pipline2}
\end{figure*}

In this paper, we focus on designing a robust and efficient instance-level CD method in 3D scenes. 
Our method aims to effectively address scenarios with uncontrolled lighting conditions while achieving fast processing times and eliminating the need for manual threshold setting. To this end, we propose \textit{Gaussian Difference} which builds on 4D Gaussian Splatting \cite{wu20244d} that extends 3D Gaussian \cite{kerbl20233d} by incorporating a deformation field to model temporal changes in Gaussian distributions. The key advantage of 4D Gaussian lies in its ability to express multi-temporal scene information within a single unified model.
We embed multiple images captured from a scene before and after the change into 4D Gaussian. Then we interpolate camera poses to generate a smooth image sequence composed of rendered images before and after the change. We segment each image with SAM \cite{kirillov2023segment} to obtain independent instances and track the instance with DEVA \cite{cheng2023tracking} to endow each instance with a unique identifier. By comparing the IDs in the image pairs, we can efficiently identify the change instances. The change maps are utilized to retrain the 4D Gaussians with newly added classification encoding, serving as a change indicator to categorize Gaussians as either changed or unchanged. Gaussian Difference can render an accurate change map of any view direction. We have conducted abundant experiments on various scenes. Compared with \text{C-NeRF} \cite{huang2023c} and CYWS-3D \cite{sachdeva2023change3D}, Gaussian Difference demonstrates comparable detection performance on the scenes with uniform lighting and significant improvements in detection accuracy on the scenes with large lighting variations. Our main contributions are as follows:
\begin{itemize}
    \item We propose Gaussian Difference, a 4D Gaussian Splatting based change detection framework, which is capable of generating a change map of any view direction.
    

    \item Gaussian Difference is efficient in training and robust to the lighting changes of scenes and does not require manually setting the threshold for each scene.
    
    %

    \item We enrich the 3D instance-level CD dataset with more scenes that have large lighting variations, which will be released to boost the study of 3D change detection.
    
    
    
\end{itemize}


\section{METHODOLOGY}
The primary objective of our research is to develop a comprehensive 3D scene representation of change instances, capable of accurately rendering images of alterations from novel viewing directions.  As illustrated in Fig.~\ref{fig:pipline2}, our 3D representations are built on 4D Gaussian splatting \cite{wu20244d}. Given two image sets $\mathcal{S}_1= \{\textbf{I}_1^1, \textbf{I}_2^1,\dots, \textbf{I}_{N_1}^1\}$ and $\mathcal{S}_2 = \{\textbf{I}_1^2, \textbf{I}_2^2,\dots, \textbf{I}_{N_2}^2\}$ captured before and after change, we initially learn a 4D Gaussian representation $\Psi$. And then we interpolate poses to generate smooth variations from $\textbf{I}_1^1$ to $\textbf{I}_{N_1}^1$ and $\textbf{I}_1^2$ to $\textbf{I}_{N_2}^2$, which can be used to generate consistent IDs of the instances across different frames. Comparing the IDs of the two scenes, we can easily obtain the change IDs and their corresponding masks with track-based instance-level change detection. Finally, we propose 4D Gaussian distribution partitioning to categorize the Gaussians into changed Gaussians $\Psi_\text{c}$ and unchanged Gaussians $\Psi_\text{uc}$, utilizing the change maps. After partitioning, we can generate a change map of any desired view direction.


\subsection{Camera pose interpolation}

Let $\mathcal{P}_1=\{p_1^1,p_2^1, \dots, p_{N_1}^1 \}$ and $\mathcal{P}_2=\{p_1^2,p_2^2, \dots, p_{N_2}^2\}$ represent the camera poses of the images in $\mathcal{S}_1$ and $\mathcal{S}_2$, respectively. 
Note that $p_i^k = (Q_i^k, T_i^k)$ denotes the camera pose of the $i$-th image of the $k$-th image set. $Q_i^k$ and $T_i^k$ denote the rotation quaternion and translation matrix, respectively.
We merge $\mathcal{P}_1$ and $\mathcal{P}_2$ to form a combined set $\mathcal{P} = \{p_1,p_2, \dots, p_{N_1}, p_{N_1+1},p_{N_1+2}, \dots, p_{N_1+N_2}\}$. 
We use spherical linear interpolation and linear interpolation to interpolate $n$ rotation quaternions and translation matrixes between two camera poses $p_i$ and $p_{i+1}$ by
\begin{equation}
    Q_{ij}= \frac{\sin((1 - \Delta_j) \theta)}{\sin(\theta)} Q_i + \frac{\sin(\Delta_j \theta)}{\sin(\theta)} Q_{i+1},
\end{equation}
\begin{equation}
    T_{ij} = (1 - \Delta_j) T_i + \Delta_j T_{i+1},
\end{equation}
where $\Delta_j = \frac{j}{n + 1}, j=1,\dots,n$,  and $\mathbf{\theta}$ is the angle between $Q_i$ and $Q_{i+1}$. Thus, we can get two novel camera poses $\mathcal{\tilde{P}}_1=\{p_1^1,p_{11}^1,\dots, p_{1n}^1,p_2^1, \dots, p_{N_1}^1 \}$ and $\mathcal{\tilde{P}}_2=\{p_1^2,p_{11}^2,\dots,p_{1n}^2,p_2^2, \dots, p_{N_2}^2\}$. 
Note that we discard the interpolation poses between $p_{N_1}$ and $p_{N_1+1}$ in $\mathcal{P}$ for later steps.
\subsection{Change detection with inconsistent IDs}
Given $\mathcal{\tilde{P}}_1$, we use the pre-trained 4D GS representation $\Psi$ to generate two image sets $\mathcal{\tilde{S}}_1 = \{\textbf{I}_1^1,\textbf{I}_{11}^1, \dots,\textbf{I}_{1n}^1,  \textbf{I}_2^1,\dots, \textbf{I}_{N_1}^1\}$ and its corresponding image set $\mathcal{\tilde{S}}_1' = \{\textbf{I}_1^2,\textbf{I}_{11}^2, \dots,\textbf{I}_{1n}^2,  \textbf{I}_2^2,\dots, \textbf{I}_{N_1}^2\}$ after change.
In order to generate a consistent ID for the same instance, we first use SAM \cite{kirillov2023segment} to segment each image to obtain the independent instances. Then, we utilize DEVA \cite{cheng2023tracking}, a pre-trained zero-shot tracker, to assign IDs to the instances. By comparing the IDs of $\text{I}_1^1$ and $\text{I}_1^2$, we identify the IDs of the change instances at the camera pose $p_1^1$. This procedure is repeated for all other poses. The masks of the changed instances serve as our change maps. Consequently, we obtain the change mask set  $\mathcal{M}_1=\{\textbf{M}_1^1,\textbf{M}_{11}^1, \dots,\textbf{M}_{1n}^1,  \textbf{M}_2^1,\dots, \textbf{M}_{N_1}^1\}$ for $\mathcal{\tilde{P}}_1$. Same operations are conducted on $\mathcal{\tilde{P}}_2$ to generate $\mathcal{M}_2=\{\textbf{M}_1^2,\textbf{M}_{11}^2, \dots,\textbf{M}_{1n}^2,  \textbf{M}_2^2,\dots, \textbf{M}_{N_2}^2\}$.

\subsection{4D Gaussian partition}
To generate change maps of new viewpoints, it is essential to identify which Gaussians correspond to the changed instances.  
Inspired by Gaussian Grouping \cite{ye2023gaussian}, we introduce classification encoding, which is a learnable vector with 16 bits. The classification of the changed instances and unchanged backgrounds should be consistent across different rendering views. The SH degree of the classification encoding is set to 0 to model its direct-current component.
For the classification encoding $e$, we adopt the same processing method as the pixel color. We calculate the 2D classification feature $f_c$ for each pixel by the weighted sum of the classification encodings that overlap with the pixel in depth order by
\begin{align}
    f_c = \sum_{i\in \mathcal{N}}e_{i}\alpha_{i}\prod_{j=1}^{i-1}(1-\alpha_{i}),
\end{align}
where $\mathcal{N}$ represents the set of Gaussian distributions that overlap with the pixels, ordered by depth, and $\alpha_i$ denotes the opacity of the $i$-th Gaussian in the sequence.
Let $\textbf{F}\in \mathbb{R}^{16\times H\times W}$ denote the classification feature matrix. Then we can generate the corresponding change map by
\begin{equation}\label{eq:cdmaps}
   \hat{\mathbf{M}} = Cov(\textbf{F}),
\end{equation}
where $Cov (\cdot)$ is a convolutional layer that transforms $\textbf{F}$ into a change mask $\hat{\mathbf{M}}$.

\subsection{Loss function}
To learn the new 4D Gaussian representation, we use
$\mathcal{S}_1= \{\textbf{I}_1^1, \textbf{I}_2^1,\dots, \textbf{I}_{N_1}^1\}$ and $\mathcal{S}_2 = \{\textbf{I}_1^2, \textbf{I}_2^2,\dots, \textbf{I}_{N_2}^2\}$ as inputs with 
$\hat{\mathcal{M}}_1=\{\textbf{M}_1^1,  \textbf{M}_2^1,\dots, \textbf{M}_{N_1}^1\}$ and $\hat{\mathcal{M}}_2=\{\textbf{M}_1^2, \textbf{M}_2^2,\dots, \textbf{M}_{N_2}^2\}$ as groundtruth change maps.
We propose a 2D loss $\mathcal{L}_{2d}$ utilizing binary cross entropy loss to classify each 3D Gaussian based on the change mask,  by
\begin{equation}
       \mathcal{L}_{2d}=-\frac{1}{\left \| \textbf{M} \right \| } \sum_{i \in \textbf{M}}\sum_{c=0}^{1}p(i)log\hat{p}(i), 
\end{equation}
where $\textbf{M}$ represents the change mask from $\mathcal{M}_1$ or $\mathcal{M}_2$.
The unsupervised 3D Regularization Loss $\mathcal{L}_{3d}$ is employed to regulate the learning of classification encoding $e_i$, which takes advantage of the 3D spatial consistency by enforcing the classification encodings of the top $k$-nearest 3D Gaussians to be close in their feature distance. $\mathcal{L}_{3d}$ is defined as
\begin{equation}
\begin{aligned}
   \mathcal{L}_{3d}&=\frac{1}{m} \sum_{m}^{j=1}D_{kl}(\mathcal{G}||\mathcal{H}) \\
   &= \frac{1}{mk}\sum_{m}^{j=1} \sum_{k}^{i=1} \mathcal{F}(e_j)log(\frac {\mathcal{F}(e_i)}{\mathcal{F}(e_j^{'})} ),
\end{aligned}
\end{equation}
where $\mathcal{G}$ consists of the sampled classification encoding $e$ of a 3D Gaussian, the set $\mathcal{H}=\{ e_1^{'},e_2^{'},\cdots,e_k^{'},\}$ represents its $k$ nearest neighbors in 3D space, and $\mathcal{F} = softmax(Cov(e_i))$.

As in the general reconstruction problem \cite{fridovich2022plenoxels, sun2022direct, fridovich2023k}, we also use L1 loss and total variation loss $\mathcal{L}_{TV}$ to supervise the quality of the rendered image. The total loss is defined as
\begin{equation}
    \mathcal{L} = \mathcal{L}_1 + \mathcal{L}_{TV} +\lambda_{2d}\mathcal{L}_{2d} + \lambda_{3d}\mathcal{L}_{3d},
\end{equation}
where $\lambda_{2d}$ and $\lambda_{3d}$ are two balance coefficients.

\subsection{Change detection in a novel view}
After training the 4D Gaussian, we can render an image $\textbf{I}^t$ as \cite{wu20244d} and generate the corresponding change map $\textbf{M}$ by Eq.\ref{eq:cdmaps} for a novel view at the pose $p$ and time $t$.

\begin{figure}[!tb]
\centering
\includegraphics[width=\linewidth]{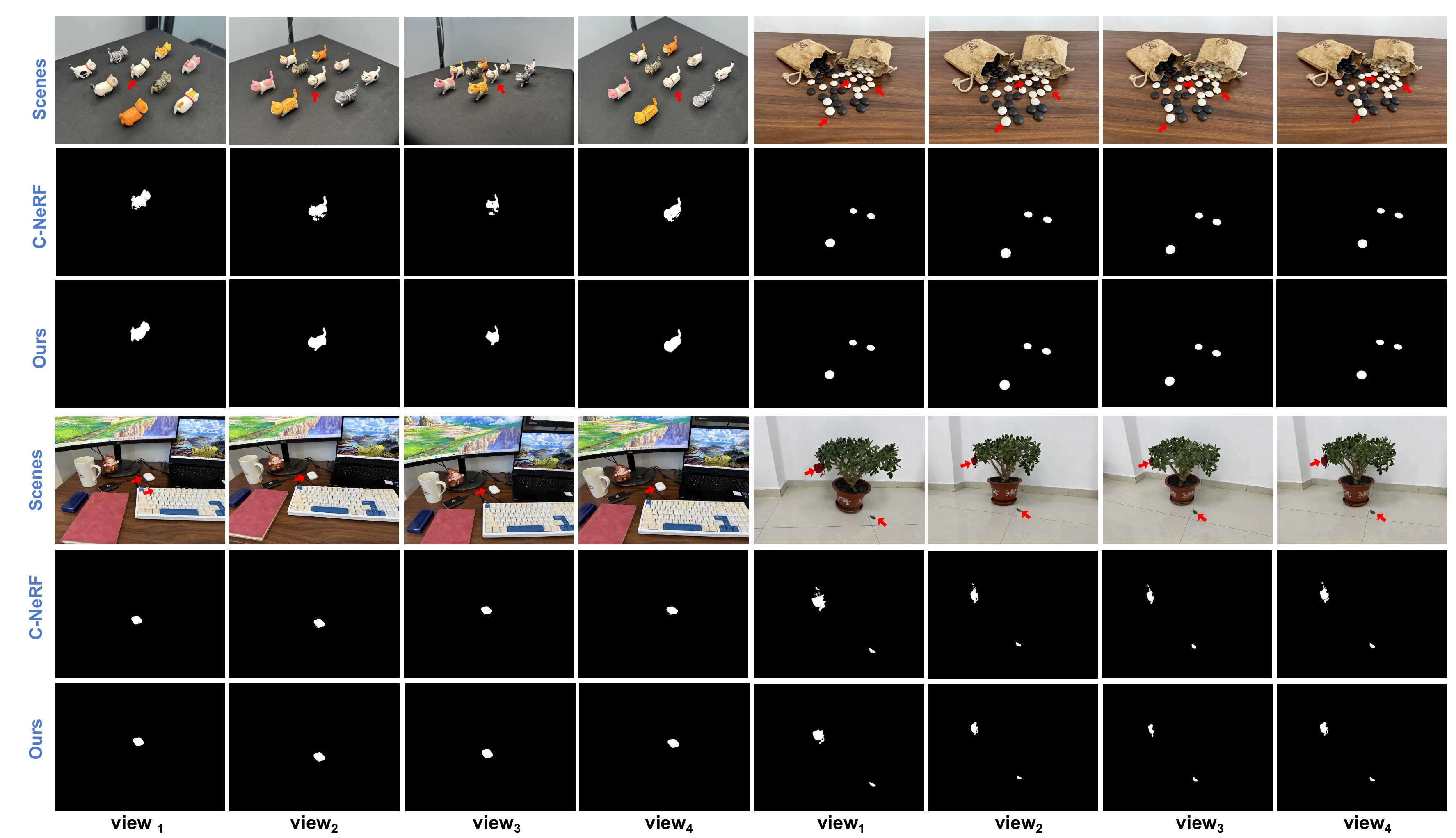}
\caption{Examples of CD results of different methods on four scenes.}
\label{fig:results1}
\end{figure}

\section{EXPERIMENT}
\subsection{Setup}
\textbf{Dataset.} We adopt four scenes of \textsc{C-NeRF} \cite{huang2023c} to evaluate the performances of different CD methods, including Desk, Chess, Potting, and Cats.
Desk consists of 40 training images, Chess has 48 training images, Potting includes 42 training images, Cats is built with 243 training images. Each scene has 10 image pairs for evaluation that are uniformly sampled from different viewpoints.


\textbf{Baselines.} \textsc{C-NeRF} \cite{huang2023c} and 
CYWS-3D \cite{sachdeva2023change3D} are used for comparison. We conduct quantitative comparisons with \textsc{C-NeRF} for it can produce pixel-level prediction. 
CYWS-3D conducts object-level change detection in two-view and class-agnostic scenarios. The input of CYWS-3D is a two-view image,  with the depth map used to obtain geometric information of the scene for change detection. The output is the bounding boxes of the changes. To compare with CYWS-3D, we use the minimum bounding rectangle of the connected area as the bounding box for our method and \textsc{C-NeRF}.

\textbf{Criteria.} We adopt Precision (P), Recall (R), F1-measure (F$_1$), and Intersection over Union (IoU) as evaluation metrics.

\subsection{Results analysis}
\begin{figure}[!tb]
  \centering
  \includegraphics[width=\linewidth]{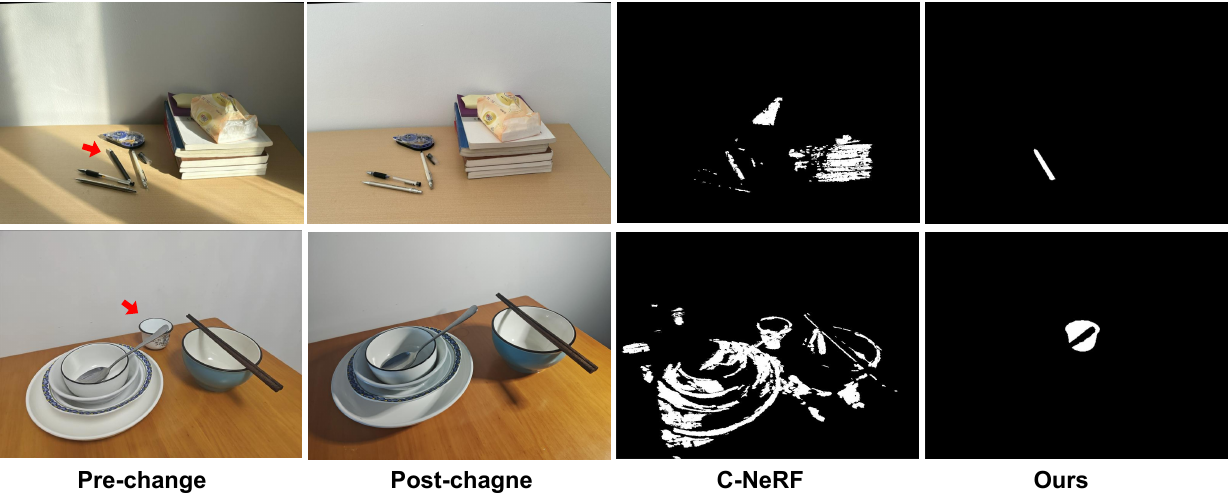}
  \caption{CD on the scenes with substantial lighting variations.} 
  \label{fig:results2}
\end{figure}

\begin{figure}[!tb]
  \centering
  \includegraphics[width=\linewidth]{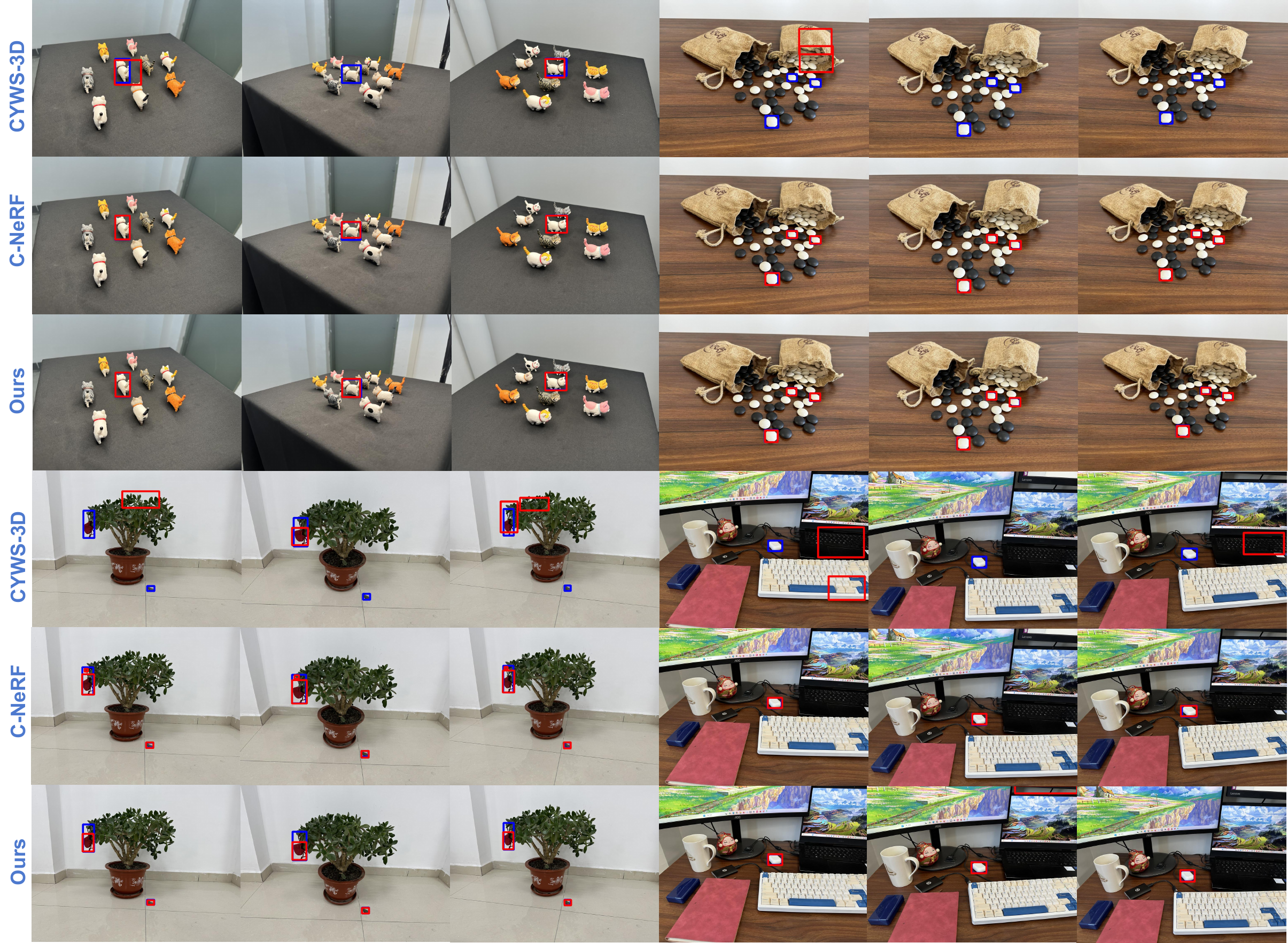}
  \caption{Comparison of CYWS-3D, \textsc{C-NeRF}, and our method on different scenes. Red boxes are the
predictions and blue boxes are the groundtruths.} 
  \label{fig:results3}
\end{figure}

Fig.~\ref{fig:results1} shows some typical detection results of our method and \textsc{C-NeRF}. Both methods successfully detect real changes across the four scenes. However, our results are more complete than the results of \textsc{C-NeRF}. In the Cats scene, the detected change region of \textsc{C-NeRF} is influenced by the shadows. To demonstrate the robustness of our method, we perform change detection on two new scenes with challenging lighting conditions. As shown in Fig.~\ref{fig:results2}, the first scene has non-uniform lighting, while the second scene has strong shadows. \textsc{C-NeRF} fails to distinguish the background change with the foreground change. However, our method can detect the changes for both scenes. In Fig.~\ref{fig:results3}, we compare the bounding box results of three CD methods. CYWS-3D can detect changes of larger objects, such as the cat (scene of Cats) and bags (scene of Potting). However, it performs poorly in detecting small objects or objects with subtle geometric changes.



The quantitative results of our method and \textsc{C-NeRF} are presented in Table.~\ref{table:res_fine_grained}. The F$_1$ and IoU values of our method are better than \textsc{C-NeRF} in three scenes. In Table.~\ref{table:running_time}, we compare the running time of the different stages on Potting. Note that our method is more efficient than \textsc{C-NeRF}.


\begin{table}[!t]
\centering
\caption{Quantitative comparison of different CD methods. The best values are highlighted in \textbf{bolded}.}
\begin{tabular}{cc|cccc}
\toprule
Scene   & Method          & P$\uparrow$ & R$\uparrow$ & F$_1$ $\uparrow$ & IoU $\uparrow$  \\ 
\midrule
\multirow{2}{*}{Desk}    & \textsc{C-NeRF}   & \textbf{96.02}       & 90.48       & \textbf{93.17}         & \textbf{87.24}   \\
                         & Ours     & 93.99       & \textbf{91.88}       & 92.78         & 86.75     \\
                        
\midrule
\multirow{2}{*}{Chess}   & \textsc{C-NeRF}   & 92.66       & \textbf{99.71}       & 96.05        & 92.41     \\
                         & Ours     & \textbf{98.64}       & 93.70       & \textbf{96.10}          & \textbf{92.51}    \\
                       
\midrule
\multirow{2}{*}{Potting} & \textsc{C-NeRF}   & 71.59       & \textbf{94.66}       & 81.75         & 69.01      \\
                         & Ours     & \textbf{93.54}       & 76.51       & \textbf{84.31}        & \textbf{72.67}      \\ 
                        
\midrule
\multirow{2}{*}{Cats} & \textsc{C-NeRF}   & 86.70       & 90.46       & 88.54         & 79.47      \\
                      & Ours     & \textbf{97.47}       & \textbf{93.21}       & \textbf{92.41}         & \textbf{91.03}      \\
                      
\bottomrule
\end{tabular}
\label{table:res_fine_grained}
\end{table}

\begin{table}[!t]
\centering
\caption{Running time of main stages of different CD methods.}
\resizebox{\linewidth}{!}{
\begin{tabular}{ccccc}
\toprule
Stage  & NeRF/4DGS & CD & Rendering/Gaussian partition  &Total\\ 
\midrule
\textsc{C-NeRF} & 16.07h & 0.75h & 0.78h &17.6h\\
Ours & 1.17h & 0.1h & 1.2h &2.47h \\
\bottomrule
\end{tabular}
}
\label{table:running_time}
\end{table}

\subsection{Discussion}
\begin{figure}[!tb]
  \centering
  \includegraphics[width=\linewidth]{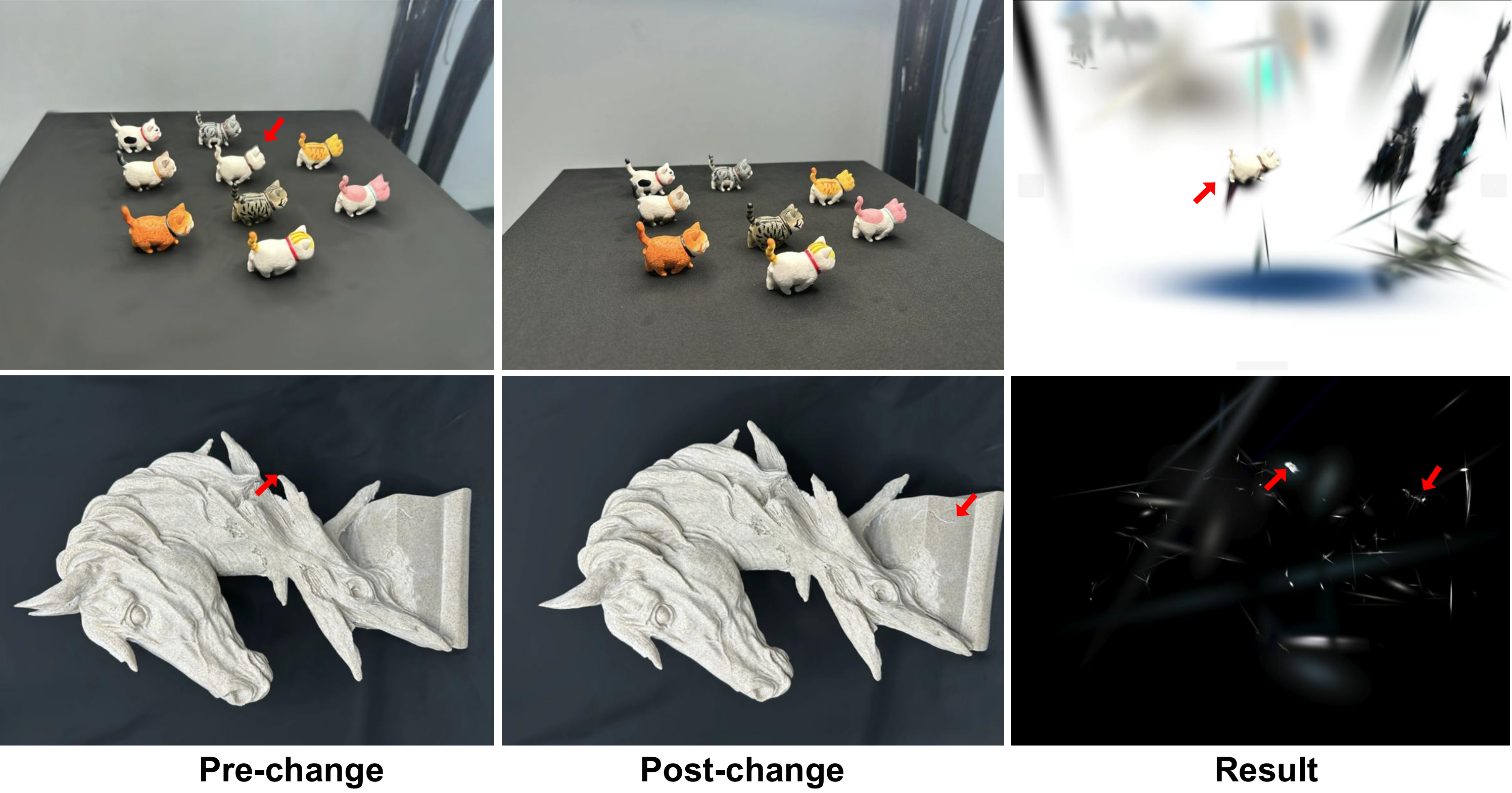}
  \caption{The CD results obtained by filtering Gaussian distributions with of $(\Delta x, \Delta r, \Delta s)$. The red arrow indicates the real change.}
  \label{fig:discussion}
\end{figure}

A 3D Gaussian is characterized by five attributes, including position $x\in \mathbb{R}^{3}$, scale factor $s\in \mathbb{R}^{3}$, rotation quaternion $r\in \mathbb{R}^{3}$, opacity $\alpha\in \mathbb{R}$, and color defined by spherical harmonic(SH) $c \in \mathbb{R}^{k}$ (where $k$ represents nums of SH functions).
In 4D Gaussian Splatting, Separate MLPs are used to compute the deformations of the position $\Delta x$, rotation $\Delta r$, and scale $\Delta s$, and represent both the Gaussian motion and deformation. A direct solution for CD is to use $(\Delta x, \Delta r, \Delta s)$ to find the changed Gaussian, and render the change map. However, as shown in Fig.~\ref{fig:discussion}, we can observe that this simple solution cannot generate the desired results. There are two main reasons: (1) Since $(\Delta x, \Delta r, \Delta s)$ change collaboratively, we cannot accurately distinguish the Gaussian into changed or unchanged;
(2) A Gaussian might be used to represent multiple objects or backgrounds. The changed Gaussian not only renders the changed objects but also renders unchanged objects or backgrounds. 

\section{Conclusion}

In this paper, we present a pioneering approach to 3D instance-level change detection, addressing the challenge of identifying changes in real-world scenarios without relying on controlled conditions or labeled image pairs. Our method employs 4D Gaussian Splatting to embed two sets of images into Gaussian representations, enabling the rendering of coherent image sequences. By segmenting images and assigning unique identifiers to each instance, we efficiently detect changes through ID comparison. We partition the Gaussians into changed and unchanged using a classification encoding for each Gaussian, with change maps generated through rendering. Extensive experiments on various datasets show that our approach significantly outperforms methods like C-NeRF and CYWS-3D, especially under varying lighting conditions.

\newpage
\bibliographystyle{unsrt}   
\bibliography{main}          

\begin{thebibliography}{10}

\bibitem{sakurada2013detecting}
Ken Sakurada, Takayuki Okatani, and Koichiro Deguchi.
\newblock Detecting changes in 3d structure of a scene from multi-view images captured by a vehicle-mounted camera.
\newblock In {\em Proceedings of the IEEE/CVF Conference on Computer Vision and Pattern Recognition}, pages 137--144, 2013.

\bibitem{alcantarilla2018street}
Pablo~F Alcantarilla, Simon Stent, German Ros, Roberto Arroyo, and Riccardo Gherardi.
\newblock Street-view change detection with deconvolutional networks.
\newblock {\em Autonomous Robots}, 42:1301--1322, 2018.

\bibitem{2d_captioning_jhamtani2018learning}
Harsh Jhamtani and Taylor Berg-Kirkpatrick.
\newblock Learning to describe differences between pairs of similar images.
\newblock In {\em Proceedings of the 2018 Conference on Empirical Methods in Natural Language Processing}, pages 4024--4034, 2018.

\bibitem{chen2020spatial}
Hao Chen and Zhenwei Shi.
\newblock A spatial-temporal attention-based method and a new dataset for remote sensing image change detection.
\newblock {\em Remote Sensing}, 12(10):1662, 2020.

\bibitem{huang2023background}
Rui Huang, Ruofei Wang, Qing Guo, Jieda Wei, Yuxiang Zhang, Wei Fan, and Yang Liu.
\newblock Background-mixed augmentation for weakly supervised change detection.
\newblock In {\em Proceedings of the AAAI Conference on Artificial Intelligence}, volume~37, pages 7919--7927, 2023.

\bibitem{lei2020hierarchical}
Yinjie Lei, Duo Peng, Pingping Zhang, Qiuhong Ke, and Haifeng Li.
\newblock Hierarchical paired channel fusion network for street scene change detection.
\newblock {\em IEEE Transactions on Image Processing}, 30:55--67, 2020.

\bibitem{wang2023cross}
Guangxing Wang, Gong Cheng, Peicheng Zhou, and Junwei Han.
\newblock Cross-level attentive feature aggregation for change detection.
\newblock {\em IEEE Transactions on Circuits and Systems for Video Technology}, 2023.

\bibitem{ji2018fully}
Shunping Ji, Shiqing Wei, and Meng Lu.
\newblock Fully convolutional networks for multisource building extraction from an open aerial and satellite imagery data set.
\newblock {\em IEEE Transactions on geoscience and remote sensing}, 57(1):574--586, 2018.

\bibitem{lebedev2018change}
MA~Lebedev, Yu~V Vizilter, OV~Vygolov, Vladimir~A Knyaz, and A~Yu Rubis.
\newblock Change detection in remote sensing images using conditional adversarial networks.
\newblock {\em The International Archives of the Photogrammetry, Remote Sensing and Spatial Information Sciences}, 42:565--571, 2018.

\bibitem{ulusoy2014image}
Ali~Osman Ulusoy and Joseph~L Mundy.
\newblock Image-based 4-d reconstruction using 3-d change detection.
\newblock In {\em Computer Vision--ECCV 2014: 13th European Conference, Zurich, Switzerland, September 6-12, 2014, Proceedings, Part III 13}, pages 31--45. Springer, 2014.

\bibitem{6942806}
Rareş Ambruş, Nils Bore, John Folkesson, and Patric Jensfelt.
\newblock Meta-rooms: Building and maintaining long term spatial models in a dynamic world.
\newblock In {\em 2014 IEEE/RSJ International Conference on Intelligent Robots and Systems}, pages 1854--1861, 2014.

\bibitem{qin20163d}
Rongjun Qin, Jiaojiao Tian, and Peter Reinartz.
\newblock 3d change detection--approaches and applications.
\newblock {\em ISPRS Journal of Photogrammetry and Remote Sensing}, 122:41--56, 2016.

\bibitem{huang2023c}
Rui Huang, Binbin Jiang, Qingyi Zhao, William Wang, Yuxiang Zhang, and Qing Guo.
\newblock C-nerf: Representing scene changes as directional consistency difference-based nerf.
\newblock {\em arXiv preprint arXiv:2312.02751}, 2023.

\bibitem{7989614}
Marius Fehr, Fadri Furrer, Ivan Dryanovski, Jürgen Sturm, Igor Gilitschenski, Roland Siegwart, and Cesar Cadena.
\newblock Tsdf-based change detection for consistent long-term dense reconstruction and dynamic object discovery.
\newblock In {\em 2017 IEEE International Conference on Robotics and Automation (ICRA)}, pages 5237--5244, 2017.

\bibitem{ku2021shrec}
Tao Ku, Sam Galanakis, Bas Boom, Remco~C Veltkamp, Darshan Bangera, Shankar Gangisetty, Nikolaos Stagakis, Gerasimos Arvanitis, and Konstantinos Moustakas.
\newblock Shrec 2021: 3d point cloud change detection for street scenes.
\newblock {\em Computers \& Graphics}, 99:192--200, 2021.

\bibitem{qiu20233d}
Yue Qiu, Shintaro Yamamoto, Ryosuke Yamada, Ryota Suzuki, Hirokatsu Kataoka, Kenji Iwata, and Yutaka Satoh.
\newblock 3d change localization and captioning from dynamic scans of indoor scenes.
\newblock In {\em Proceedings of the IEEE/CVF Winter Conference on Applications of Computer Vision}, pages 1176--1185, 2023.

\bibitem{sachdeva2023change2D}
Ragav Sachdeva and Andrew Zisserman.
\newblock The change you want to see.
\newblock In {\em Proceedings of the IEEE/CVF Winter Conference on Applications of Computer Vision}, pages 3993--4002, 2023.

\bibitem{mildenhall2020nerf}
Ben Mildenhall, Pratul~P. Srinivasan, Matthew Tancik, Jonathan~T. Barron, Ravi Ramamoorthi, and Ren Ng.
\newblock Nerf: Representing scenes as neural radiance fields for view synthesis.
\newblock In {\em ECCV}, 2020.

\bibitem{wu20244d}
Guanjun Wu, Taoran Yi, Jiemin Fang, Lingxi Xie, Xiaopeng Zhang, Wei Wei, Wenyu Liu, Qi~Tian, and Xinggang Wang.
\newblock 4d gaussian splatting for real-time dynamic scene rendering.
\newblock In {\em Proceedings of the IEEE/CVF Conference on Computer Vision and Pattern Recognition}, pages 20310--20320, 2024.

\bibitem{kerbl20233d}
Bernhard Kerbl, Georgios Kopanas, Thomas Leimk{\"u}hler, and George Drettakis.
\newblock 3d gaussian splatting for real-time radiance field rendering.
\newblock {\em ACM Trans. Graph.}, 42(4):139--1, 2023.

\bibitem{kirillov2023segment}
Alexander Kirillov, Eric Mintun, Nikhila Ravi, Hanzi Mao, Chloe Rolland, Laura Gustafson, Tete Xiao, Spencer Whitehead, Alexander~C Berg, Wan-Yen Lo, et~al.
\newblock Segment anything.
\newblock In {\em Proceedings of the IEEE/CVF International Conference on Computer Vision}, pages 4015--4026, 2023.

\bibitem{cheng2023tracking}
Ho~Kei Cheng, Seoung~Wug Oh, Brian Price, Alexander Schwing, and Joon-Young Lee.
\newblock Tracking anything with decoupled video segmentation.
\newblock In {\em Proceedings of the IEEE/CVF International Conference on Computer Vision}, pages 1316--1326, 2023.

\bibitem{sachdeva2023change3D}
Ragav Sachdeva and Andrew Zisserman.
\newblock The change you want to see (now in 3d).
\newblock In {\em Proceedings of the IEEE/CVF International Conference on Computer Vision}, pages 2060--2069, 2023.

\bibitem{ye2023gaussian}
Mingqiao Ye, Martin Danelljan, Fisher Yu, and Lei Ke.
\newblock Gaussian grouping: Segment and edit anything in 3d scenes.
\newblock {\em arXiv preprint arXiv:2312.00732}, 2023.

\bibitem{fridovich2022plenoxels}
Sara Fridovich-Keil, Alex Yu, Matthew Tancik, Qinhong Chen, Benjamin Recht, and Angjoo Kanazawa.
\newblock Plenoxels: Radiance fields without neural networks.
\newblock In {\em Proceedings of the IEEE/CVF conference on computer vision and pattern recognition}, pages 5501--5510, 2022.

\bibitem{sun2022direct}
Cheng Sun, Min Sun, and Hwann-Tzong Chen.
\newblock Direct voxel grid optimization: Super-fast convergence for radiance fields reconstruction.
\newblock In {\em Proceedings of the IEEE/CVF conference on computer vision and pattern recognition}, pages 5459--5469, 2022.

\bibitem{fridovich2023k}
Sara Fridovich-Keil, Giacomo Meanti, Frederik~Rahb{\ae}k Warburg, Benjamin Recht, and Angjoo Kanazawa.
\newblock K-planes: Explicit radiance fields in space, time, and appearance.
\newblock In {\em Proceedings of the IEEE/CVF Conference on Computer Vision and Pattern Recognition}, pages 12479--12488, 2023.

\end{thebibliography}
\end{document}